%% file: main.tex
\documentclass[letterpaper,twocolumn,10pt]{article}
\usepackage{usenix2019_v3}

\usepackage{amsmath,amssymb,amsfonts}
\usepackage{graphicx}
\usepackage{booktabs}
\usepackage{float}
\usepackage{listings}
\usepackage{subcaption}
\usepackage{xspace}
\usepackage{tikz}
\usetikzlibrary{arrows.meta, positioning, calc, shapes.geometric, fit, shadows}

\AtBeginDocument{\DeclareMathAlphabet{\mathcal}{OMS}{cmsy}{m}{n}}

\newcommand{\openkedge}{\mbox{OpenKedge}\xspace}

\definecolor{slate}{RGB}{112,128,144}
\definecolor{emerald}{RGB}{80,200,120}

\begin{document}

\title{\bf Semantic Quorum Assurance: Collective Certification for Non-Deterministic AI Infrastructure}

% Author block - anonymous for review, or openkedge author format
\author{
  {\rm Jun He}\\
  OpenKedge.io
  \and
  {\rm Deying Yu}\\
  OpenKedge.io
}

\maketitle

\input{sections/01-introduction}
\input{sections/02-motivation}
\input{sections/03-background}
\input{sections/04-system-model}
\input{sections/05-threat-model}
\input{sections/06-semantic-quorum-assurance}
\input{sections/07-protocol-design}

\input{sections/08-sovereign-execution-gating}
\input{sections/09-implementation}
\input{sections/10-evaluation}
\input{sections/11-discussion}
\input{sections/12-related-work}
\input{sections/13-limitations}
\input{sections/13-conclusion}

\appendix
\input{sections/appendix-scenario-provenance}

\bibliographystyle{unsrt}
\bibliography{refs}

\end{document}

%% file: sections/01-introduction.tex
\begin{abstract}
As large language model (LLM) agents are integrated into autonomous cloud operations, distributed systems face a semantic reliability problem: proposer agents can generate production mutations, such as modifying IAM policies, opening firewall security groups, or executing data exports, that are syntactically valid and statically authorized but operationally unsafe. Classical distributed consensus protocols replicate deterministic state transitions but do not evaluate the safety of the proposed intent. To address this gap, we introduce Semantic Quorum Assurance (SQA), a control-plane primitive for governing non-deterministic agentic infrastructure. SQA represents proposals as declarative execution contracts bound to cryptographic evidence chains and routes them to a diverse panel of read-only, sandboxed validator agents. SQA aggregates their judgments under a risk-adaptive quorum predicate that enforces model and archetype diversity, adjusts weights based on calibrated assurance scores, and respects archetype-specific vetoes. Admitted proposals execute only through a sovereign execution gate. We instantiate SQA in a cloud-native control plane and formalize a correlated cognitive failure model for non-deterministic validators. On 500 infrastructure-inspired mutation scenarios, with safety results reported on held-out safe/unsafe trials excluding ambiguous scenarios, SQA reduces unsafe approval from 18.5\% for single-agent validation to 0.3\% while adding median validation latency of 1.45--4.12 seconds across the studied risk buckets.
\end{abstract}

\section{Introduction}
\label{sec:intro}

Autonomous agents are acquiring mutative authority over cloud control planes. Using tool-enabled language models, these agents propose and apply infrastructure changes, call Kubernetes and cloud APIs, modify access-control configurations, and remediate incidents~\cite{yao-react, schick-toolformer}. This shift introduces a systems problem: a non-deterministic reasoning process becomes the proximate cause of an irreversible state transition.

Current defenses authorize access and ensure consistency, but do not evaluate semantic safety. IAM roles, service control policies, and admission webhooks restrict access and reject structurally invalid requests. Classical consensus and state machine replication protocols order and replicate transitions, ensuring nodes apply them consistently~\cite{lamport-paxos,ongaro-raft,castro-pbft,schneider-smr}. Neither class certifies whether the proposed intent is operationally safe. A request can be authorized, syntactically correct, and consistently replicated, yet still be catastrophic in context.

Consider an autonomous cloud operations agent, $\mathcal{A}_{\text{ops}}$, asked to resolve a localized database connection timeout. The agent infers that a firewall rule is blocking traffic and generates a Terraform mutation that opens database ingress to the public Internet:
\begin{verbatim}
ingress {
  from_port   = 0
  to_port     = 0
  protocol    = "-1"
  cidr_blocks = ["0.0.0.0/0"]
}
\end{verbatim}
The mutation satisfies syntax checks and runs under an authorized deployment role. Structural policy engines confirm that the agent can manage security groups, but lack the incident context to recognize that public ingress is a disproportionate response to a routing issue. Audit logs record the decision post-facto. Human review adds latency and scales poorly in high-volume control planes. While multi-agent debate provides critique, it does not produce a gate-verifiable artifact that a control plane can enforce.

To address these limitations, we propose \textbf{Semantic Quorum Assurance (SQA)}, a control-plane primitive that introduces a semantic admission layer for non-deterministic agentic operations. SQA separates the replication of transitions (handled by consensus protocols) from the certification of transition safety. By routing proposals through a validation quorum before they reach the execution gate, SQA prevents unsafe operations from mutating the system state.

SQA decouples semantic validation from execution authority. Proposers express mutations as an \textit{execution contract} $C$ and bind them to an \textit{evidence chain} $E$ containing the observations and tools used to justify the action. The control plane routes $(C,E)$ to a heterogeneous quorum of sandboxed, read-only validators with distinct model families and archetypes. Validators emit signed vote records binding their decision, calibrated assurance, and rationale to the contract hash, evidence hash, and active sequence epoch. An aggregator evaluates these votes via a risk-adaptive quorum predicate that accounts for consequence scores, confidence calibration, archetype weights, vetoes, and estimated correlation. The sovereign execution gate commits the mutation only after verifying the selected quorum, signatures, and predicate configuration.

The objective is not to prove absolute semantic safety, but to cryptographically enforce a calibrated, risk-adaptive certification process before state changes. On infrastructure-inspired scenarios, SQA reduces unsafe approvals relative to single-agent validation, static policy, and homogeneous quorums, while preserving a mechanically checkable gate-integrity property.

This paper makes six contributions:
\begin{itemize}
    \item \textbf{Primitive and system model.} We define SQA as a control-plane primitive for certifying agentic mutations before execution, using execution contracts, evidence chains, validator registries, signed votes, and sequence-epoch-bound proofs.
    \item \textbf{Correlated cognitive failure model.} We formalize validator failure as a semantic, domain-dependent event, modeling the correlations that arise from shared model families, prompts, and training data.
    \item \textbf{Risk-adaptive quorum predicate.} We design a quorum predicate that scales validator selection and admission thresholds based on consequence scores, confidence calibration, and estimated correlation.
    \item \textbf{Sovereign gate architecture.} We design an execution gate that separates validation from mutation authority, verifying the selected quorum, votes, and evidence signatures before committing state.
    \item \textbf{Prototype implementation.} We implement a cloud-native prototype with sandboxed validators, a Go-based contract router and aggregator, BLS aggregate signatures, and a Kubernetes admission-gate integration.
    \item \textbf{Evaluation.} We evaluate SQA in the prototype on 500 infrastructure-inspired Kubernetes, database, and IAM/security scenarios. SQA reduces the unsafe approval rate from 18.5\% to 0.3\% on held-out trials, with a median validation latency of 1.45--4.12 seconds.
\end{itemize}

%% file: sections/02-motivation.tex
\section{Motivation}
\label{sec:motivation}

As cloud complexity grows, organizations are replacing static scripts with autonomous, agentic infrastructure. These systems employ LLM-based agents for open-ended tasks like performance tuning, vulnerability remediation, and cost optimization. This transition from deterministic code to model-driven reasoning exposes control planes to semantic failure modes that conventional checks cannot detect.

\subsection{Agentic Failure Modes}
Traditional software typically fails in predictable, syntax-driven ways, such as null-pointer dereferences or regular expression mismatches. In contrast, LLM agents fail semantically:
\begin{itemize}
    \item \textbf{Reasoning Hallucinations:} The agent executes a logical plan based on incorrect associations, such as confusing production and staging environments due to naming similarities.
    \item \textbf{Stale Evidence Bias:} The agent reads cached or outdated metrics and triggers mutations that are inappropriate for the active system state.
    \item \textbf{Prompt Drift \& Injection:} Changes in user prompts or untrusted text in system logs cause the agent to deviate from its safety constraints.
\end{itemize}

\subsection{Illustrative Scenario: Backup Partition Cleanup}
Consider an autonomous database administration agent, $\mathcal{A}_{\text{dba}}$, tasked with maintaining database disk utilization below 80\% in a Kubernetes cluster. During a storage alert, $\mathcal{A}_{\text{dba}}$ diagnoses the primary database and finds that the historical backup volume, \texttt{backup-db-vol-01}, is at 95\% capacity. The agent reasons that:
\begin{enumerate}
    \item Old backup files can be safely pruned to free up storage space.
    \item The volume contains a directory named \texttt{/backups/active/}.
    \item The agent mistakes the word \texttt{active} for ``currently compiling backups'' rather than ``active production restore points'' due to an ambiguous context window.
\end{enumerate}

To free space, the agent decides to delete the contents of \texttt{/backups/active/}. It invokes its authorized Kubernetes tool attachment to run:
\begin{center}
\footnotesize
\texttt{kubectl exec -it db-client-pod \textbackslash} \\
\texttt{~~-- rm -rf /backups/active/*}
\end{center}
This command is syntactically correct, matches authorized pattern regexes, and executes under sufficient Kubernetes RBAC permissions. However, it is semantically catastrophic: it deletes the database's only active restore points, exposing the system to permanent data loss during a disk failure.

\subsection{Why Existing Solutions Fail}
Existing protection mechanisms do not detect this semantic failure at admission time.

\subsubsection{Why Classical Consensus is Insufficient}
Classical consensus protocols (e.g., Paxos~\cite{lamport-paxos}, Raft~\cite{ongaro-raft}, PBFT~\cite{castro-pbft}) solve state agreement under crash or Byzantine faults. They enforce state machine replication (SMR)~\cite{schneider-smr} to ensure replicas execute identical state transitions. However, they assume:
\begin{enumerate}
    \item State transitions are deterministic.
    \item Proposed transitions are validated by strict, deterministic rules.
\end{enumerate}

In agentic systems, failures are semantic rather than physical or network-related. If an agent proposes \texttt{rm -rf /backups/active/*}, consensus protocols will faithfully replicate and commit the command across all nodes, destroying the data. Classical consensus guarantees consistent replication, not semantic safety.

\subsubsection{Static Policy Engines}
Policy-as-code engines (e.g., Open Policy Agent) evaluate requests against static conditions, such as blocking database mutations outside maintenance windows. They cannot evaluate natural-language context or verify the agent's reasoning. If a command satisfies structural rules, policy engines permit it regardless of operational safety.

\subsubsection{Audit Logging}
Audit logs (e.g., AWS CloudTrail) record events retrospectively and provide no prevention. By the time an operator is alerted to backup deletion, the data is gone.

\subsubsection{Human-in-the-Loop}
Manual human review creates a bottleneck that defeats the speed and automation benefits of agentic control. Human operators also face alert fatigue: reviewing dozens of benign agent requests per hour leads to careless approvals.

\subsection{The SQA Approach}
SQA introduces a semantic validation quorum before execution. Instead of relying solely on replication or static checks, SQA routes agent proposals to a diverse quorum of validator agents. These validators review both the agent's rationale and the execution details. The control plane admits the transition only when the quorum emits a risk-adaptive certificate verifiable by the execution gate.

%% file: sections/03-background.tex
\section{Background and Context}
\label{sec:background}

This section summarizes the OpenKedge control-plane primitives that SQA builds upon: bounded execution contracts, reasoning-execution separation, and cryptographic evidence records.

\subsection{Intent-Based Bounded Mutation}
Under a governed agentic control plane, agents cannot access open-ended terminals. Instead, they propose actions within a bounded schema called an \textit{execution contract}. The contract specifies:
\begin{itemize}
    \item The target resources to be modified.
    \item The exact operations to be performed (e.g., restart, prune, scale).
    \item The system invariants that must be maintained post-mutation.
\end{itemize}
This bounds the agent's action space and prevents command execution from escaping structural parameters.

\subsection{Sovereign Agentic Loops (SAL)}
The SAL model separates reasoning from execution. Non-deterministic reasoning runs in a multi-tenant cloud environment (the \textit{foreign reasoning engine}), which outputs a proposed contract. Contract execution is performed locally by a trusted component (the \textit{sovereign execution gate}) within the secure operational boundary. This separation ensures that reasoning models lack direct access to managed systems.

\subsection{Verifiable Agentic Infrastructure (VAI)}
Under VAI, each agent has a cryptographic execution identity. As the agent observes the system and executes tools, the control plane records these interactions in an \textit{evidence chain}. The evidence chain $E$ is an append-only, signed ledger of:
\begin{itemize}
    \item The system observations (metrics, logs) read by the agent.
    \item The tool outputs (CLI results, API calls) retrieved by the agent.
    \item Structured rationale artifacts or reasoning summaries generated by the model.
\end{itemize}
This chain provides a verifiable audit trail of the agent's decision process.

\subsection{The Role of SQA in OpenKedge}
SQA provides the semantic admission layer for OpenKedge. While the execution contract restricts the scope of action, and the SAL/VAI layers verify identity and provenance, the control plane still requires semantic safety evaluation. SQA intercepts the contract and evidence chain, routes them to a validator quorum, and generates a quorum proof for the execution gate. This ensures collective verification before state changes.

%% file: sections/04-system-model.tex
\section{System Model}
\label{sec:system-model}

This section formalizes the system entities, validator agents, evaluation functions, and failure correlations within the control plane.

\subsection{System Entities}
Let $S_{\mathit{seq}} \in \mathcal{S}$ denote the state of the infrastructure at sequence epoch $\mathit{seq} \in \mathbb{N}$, where $\mathcal{S}$ represents the set of all possible configurations of resources (e.g., compute instances, network security group rules, access control policies) managed by a trusted sovereign control plane $G$. The sequence epoch is a monotonically increasing counter maintained by the control plane to track state transitions. While a single $\mathit{seq}$ implies a global ledger, practical implementations partition this sequence by resource scope to prevent concurrency bottlenecks (discussed further in Section 11.5). The core entities of SQA are formalized as follows:

\begin{itemize}
    \item \textbf{Proposed Mutation ($\Delta S_C$):} A declarative set of operations (creation, modification, deletion) targeting specific resources, representing a transition from state $S_{\mathit{seq}}$ to $S_{\mathit{seq}+1}$.
    
    \item \textbf{Execution Contract ($C$):} A formal request containing the proposed mutation and a set of declarative invariants $I_C$ that must be preserved post-execution:
    \begin{equation}
        C = (\Delta S_C, I_C)
    \end{equation}
    
    \item \textbf{Evidence Chain ($E$):} A cryptographically chained, append-only ledger compiled by the proposer agent $p$ during its reasoning cycle:
    \begin{equation}
        E = (o_1, t_1, r_1, \dots, o_k, t_k, r_k)
    \end{equation}
    where $o_j$ represents observations of state $S_{\mathit{seq}}$, $t_j$ represents tools executed (e.g., read-only CLI commands), and $r_j$ represents structured rationale artifacts or reasoning summaries from the proposer.
    
    \item \textbf{Validator Registry ($V$):} A registry of specialized, non-deterministic validator agents managed by the control plane:
    \begin{equation}
        V = \{a_1, a_2, \dots, a_N\}
    \end{equation}
    
    \item \textbf{Quorum ($Q$):} A subset of selected validators assigned to evaluate contract $C$, where $Q \subseteq V$.
\end{itemize}

\subsection{Validator Agent Model}
A validator agent $a_i \in V$ is represented by a tuple of its identity, reasoning engine, retrieval bounds, archetype, authority weight, and public key:
\begin{equation}
    a_i = (\mathit{validator\_id}_i, M_i, R_i, \kappa_i, w_i, P_i)
\end{equation}
where:
\begin{itemize}
    \item $\mathit{validator\_id}_i$ is the stable registry identity of validator $a_i$.
    \item $M_i \in \mathcal{M}$ represents the model family or inference backend. The model family determines the base architecture, pre-training corpus, serving stack, and fine-tuning parameters of the validator.
    \item $R_i \subseteq \mathcal{U}$ defines the read-only contextual retrieval capability of the validator, where $\mathcal{U}$ is the set of authorized system telemetry tables, logs, and configuration databases.
    \item $\kappa_i \in \mathcal{K}$ is the institutional archetype of the validator, such as a Security Auditor, Database Admin, Compliance Officer, or Cost Optimizer. The archetype defines the system prompt constraints, local invariants, and evaluation guidelines.
    \item $w_i \in \mathbb{R}^+$ is the domain authority weight assigned to the validator. The weight represents the validator's historical reliability and expertise in the domain of contract $C$.
    \item $P_i$ is the validator's registered public verification key.
\end{itemize}

\subsection{Semantic Evaluation Function}
Upon submission of $C$ and $E$, each selected validator $a_i \in Q$ executes a semantic evaluation function within an ephemeral sandbox:
\begin{equation}
    \sigma_i(C, E, S_{\mathit{seq}} \restriction_{R_i}) \rightarrow (v_i, c_i, r_i, \pi_i)
\end{equation}
where:
\begin{itemize}
    \item $S_{\mathit{seq}} \restriction_{R_i}$ is the projection of the global state $S_{\mathit{seq}}$ restricted to the authorized read set $R_i$, preventing the validator from accessing unauthorized system state. To preserve network isolation (i.e., \texttt{--net=none} sandboxing), this state projection is pre-fetched by the trusted control plane $G$ prior to sandbox initialization and mounted directly into the validator's isolated filesystem as a read-only file.
    \item $v_i \in \{-1, +1\}$ is the categorical vote, where $+1$ represents approval and $-1$ represents rejection.
    \item $c_i \in (0, 1]$ is the calibrated assurance score representing the quantified certainty in the vote. The score $c_i$ is not the raw self-confidence of the model. It is computed and calibrated by the control plane based on historical validator reliability, archetype-specific performance, evidence consistency, and optionally model uncertainty signals.
    \item $r_i$ is a structured rationale or evidence reference explaining the reasoning behind the vote.
    \item $\pi_i$ is a cryptographic signature generated using the validator's private key. Let $\mathit{archetype}_i$ denote the canonical serialized identifier for $\kappa_i$. The signed vote record is:
    \begin{equation}
        \begin{aligned}
        \textsf{VoteRecord}_i = (&v_i, c_i, r_i, h(C), h(E), \mathit{seq},\\
        &\mathit{validator\_id}_i, \mathit{archetype}_i)
        \end{aligned}
    \end{equation}
    The signature binds exactly this record:
    \begin{equation}
        \pi_i = \text{Sign}_{K_i}(\textsf{VoteRecord}_i)
    \end{equation}
\end{itemize}

\subsection{Correlated Cognitive Failure}
\label{subsec:correlated-failure}
To capture validator unreliability, let $\mathcal{C}_D$ be the contract distribution for domain $D$. For a contract $C \sim \mathcal{C}_D$, let $F_i(C) \in \{0, 1\}$ indicate whether validator $a_i$ fails:
\begin{equation}
    F_i(C) = 
    \begin{cases} 
      1 & \text{if } a_i \text{ incorrectly evaluates contract } C \\
      0 & \text{otherwise}
    \end{cases}
\end{equation}
Unlike classical systems, LLM validator failures are correlated by shared pre-training data, model architectures, prompts, and common biases. The correlation coefficient $\rho_{ij}(D)$ between validators $a_i$ and $a_j$ in domain $D$ is:
\begin{equation}
    \rho_{ij}(D) = \frac{\text{Cov}(F_i(C), F_j(C))}{\sqrt{\text{Var}(F_i(C))\text{Var}(F_j(C))}}
\end{equation}
where the expectations, covariance, and variances are taken over the contract distribution $C \sim \mathcal{C}_D$. The correlation $\rho_{ij}(D)$ is estimated empirically from historical validation logs, adjudicated tasks, red-teaming datasets, or historical disagreements within domain $D$. Given a calibration set $\mathcal{C}_{D,\text{cal}}$, we use the estimator:
\begin{equation}
    \hat{\rho}_{ij}(D) =
    \frac{\sum_{C \in \mathcal{C}_{D,\text{cal}}} \Delta F_i(C)\Delta F_j(C)}
    {\sqrt{\sum_{C \in \mathcal{C}_{D,\text{cal}}} \Delta F_i(C)^2 \sum_{C \in \mathcal{C}_{D,\text{cal}}} \Delta F_j(C)^2}}
\end{equation}
where $\Delta F_i(C) = F_i(C) - \mu_i$ and $\mu_i$ is the empirical failure rate of validator $a_i$ over $\mathcal{C}_{D,\text{cal}}$. When validators share a homogeneous model family and no stronger calibration evidence is available, the control plane conservatively treats $\rho_{ij}(D)$ as high, up to the fully correlated case $\rho_{ij}(D)=1.0$. Enforcing a diversity constraint does not guarantee the complete elimination of failures, but it bounds the correlated-failure risk under the estimated joint distribution.

\subsection{Notation Summary}
\begin{table}[H]
  \centering
  \scriptsize
  \begin{tabular}{@{}p{0.31\columnwidth}p{0.63\columnwidth}@{}}
    \toprule
    \textbf{Symbol} & \textbf{Meaning} \\
    \midrule
    $S_{\mathit{seq}}$ & Infrastructure state at sequence epoch $\mathit{seq}$ \\
    $\Delta S_C$ & Proposed mutation carried by contract $C$ \\
    $I_C$ & Declarative invariants required by contract $C$ \\
    $C=(\Delta S_C,I_C)$ & Execution contract \\
    $E$ & Evidence chain bound to $C$ by digest $h(E)$ \\
    $V$ & Validator registry \\
    $a_i$ & Validator agent in $V$ \\
    $\mathit{validator\_id}_i$ & Stable identity of validator $a_i$ \\
    $\kappa_i$ / $\mathit{archetype}_i$ & Validator archetype and its serialized record field \\
    $Q$ & Selected validator quorum, $Q \subseteq V$ \\
    $H(C,S_{\mathit{seq}},\cdot)$ & Deterministic constrained quorum-selection algorithm over the registry snapshot \\
    $\textsf{VoteRecord}_i$ & Canonical signed vote record for validator $a_i$ \\
    $\Gamma$ & Quorum proof presented to the execution gate \\
    $\sigma_{\text{agg}}$ & Aggregate signature over distinct signed vote records \\
    $\mathit{registry\_snapshot}$ & Immutable registry snapshot used by quorum selection \\
    $\mathsf{pred\_cfg}$ & Predicate version, thresholds, diversity bounds, and veto configuration \\
    $D$ & Operational domain of contract distribution $\mathcal{C}_D$ \\
    $R(C)$, $R_{\text{base}}(C)$ & Consequence score and base consequence score \\
    $B(C)$, $P(C)$, $I(C)$ & Blast radius, privilege level, and irreversibility scores \\
    $\textsf{Data}(C)$, $U(C)$ & Data sensitivity and telemetry uncertainty scores \\
    $\gamma,\lambda,\eta$ & Risk scaling parameters configured by the control plane \\
    $\rho_{ij}(D)$ & Empirical cognitive-failure correlation for validators $a_i,a_j$ in domain $D$ \\
    $\epsilon(R(C))$, $\tau(R(C))$ & Risk-adaptive correlation and weighted-approval thresholds \\
    $\beta_M,\beta_\kappa$ & Model-family and archetype diversity bounds \\
    $\mathcal{K}_{\text{crit}}(C)$ & Veto-enabled archetypes required for contract $C$ \\
    \bottomrule
  \end{tabular}
  \caption{Compact notation summary for SQA.}
  \label{tab:notation}
\end{table}

%% file: sections/05-threat-model.tex
\section{Threat Model}
\label{sec:threat-model}

This threat model captures both traditional security risks and the semantic failure modes of non-deterministic AI agents.

\subsection{Security Assumptions}
We assume a trusted host (optionally TEE-backed) runs the sovereign execution gate. Cryptographic keys and root credentials remain secure. However, proposers and validators may execute erratically, fail to reason correctly, or be compromised.

\subsection{Adversary Model and Attack Vectors}
The adversary aims to commit an unsafe mutation $\Delta S_C$ that violates safety invariants. Attack vectors include:
\begin{itemize}
    \item \textbf{Proposer Compromise:} An adversary gains access to a proposer agent $\mathcal{A}_{\text{prop}}$ or manipulates its inputs to propose a malicious contract $C$.
    \item \textbf{Prompt Injection:} The adversary injects instructions into the system environment (e.g., in a database log or application error message) that the proposer reads, causing it to generate a harmful contract under the guise of an operational fix.
    \item \textbf{State Tampering:} The adversary manipulates external system metrics to feed stale or incorrect evidence $E$ to the validators, inducing a false approval.
    \item \textbf{Validator Indirect Prompt Injection (IPI):} The adversary embeds malicious instructions inside the execution contract $C$, its parameters, or the reasoning chain elements of the evidence chain $E$. Because validators must parse these inputs to evaluate safety, the embedded instructions can exploit the validator models' instruction-following behavior, hijacking their judgment to force an approval.
\end{itemize}

\subsection{Correlated Cognitive Failure Model}
Classical Byzantine Fault Tolerance (BFT) assumes independent node failures. Under BFT, a system of $N \ge 3f + 1$ nodes tolerates up to $f$ hardware or network faults. This independence assumption is unsafe for language model validators. If multiple validators share a model family $M_i$, a training-data bias or prompt weakness can cause them to fail identically. Their failures are \textbf{semantically correlated}.

We use the formal correlated-failure model defined in Section~\ref{subsec:correlated-failure}. In that model, $F_i(C)$ records whether validator $a_i$ incorrectly evaluates contract $C$, and $\rho_{ij}(D)$ measures empirical failure correlation between validators $a_i$ and $a_j$ over domain $D$.

When validators share the same model family, prompt template, or retrieval corpus, the threat model treats $\rho_{ij}(D)$ as potentially high unless calibration data justifies a lower estimate. Even when different model families are used, they may share common pre-training sources, benchmark exposure, or prompt-following biases, causing $\rho_{ij}(D)$ to remain above $0$.

Therefore, standard threshold majorities (e.g., simple majority voting) are insufficient when validator failures are correlated. SQA explicitly accounts for $\rho_{ij}(D)$ in its quorum selection and validation predicate, enforcing model and archetype diversity to reduce estimated joint failure risk. Furthermore, if a validator is actively compromised (e.g., via Indirect Prompt Injection) and intentionally falsifies a high assurance score to push an unsafe approval, SQA limits its impact through the sovereign gate's strict verification of the exact selected quorum and the domain authority weight ($w_i$), preventing a single compromised actor from satisfying the risk-adaptive threshold alone.

%% file: sections/06-semantic-quorum-assurance.tex
\section{Semantic Quorum Assurance}
\label{sec:sqa}

At the core of Semantic Quorum Assurance is the quorum predicate $\Pi(C, Q, \mathit{seq})$, which defines the conditions under which a proposed contract $C$ is admissible by a selected validator quorum $Q \subseteq V$ at sequence epoch $\mathit{seq}$. The predicate is evaluated under an explicit predicate version and configuration, denoted $\mathsf{pred\_cfg}$, and this configuration is recorded in the quorum proof.

\subsection{Consequence Score}
Infrastructure mutations carry varying levels of operational and security risk. SQA computes a consequence score, $R(C) \in [0, 1]$, for each contract $C$. Rather than averaging threat vectors, SQA models risk non-linearly using the maximum exposure across independent dimensions. The score combines a base risk score $R_{\text{base}}(C) \in [0, 1]$ with telemetry uncertainty $U(C) \in [0, 1]$:
\begin{equation}
    R(C) = 1 - (1 - R_{\text{base}}(C))(1 - \eta U(C))
\end{equation}
where:
\begin{equation}
    R_{\text{base}}(C) = \max( B(C) \cdot P(C), \gamma I(C), \lambda\,\textsf{Data}(C) )
\end{equation}
and:
\begin{itemize}
    \item $B(C) \in [0, 1]$ is the blast radius, reflecting the volume or fraction of infrastructure resources affected by the mutation.
    \item $P(C) \in [0, 1]$ is the privilege level, representing the administrative privilege or credential authority required to execute the mutation.
    \item $I(C) \in [0, 1]$ is the irreversibility score (e.g., $1.0$ for raw file deletion or database drops, $0.0$ for non-disruptive parameter updates).
    \item $\textsf{Data}(C) \in [0, 1]$ is the data sensitivity score, reflecting the confidentiality tier of the resources touched.
    \item $U(C) \in [0, 1]$ is the uncertainty score, which increases when the evidence chain $E$ is short, stale, or lacks necessary telemetry.
    \item $\gamma, \lambda, \eta \in [0, 1]$ are scaling parameters configured by the control plane.
\end{itemize}
This formulation has two key benefits. First, it reflects the fact that infrastructure risk is determined by maximum exposure rather than average risk (e.g., an irreversible deletion is high-risk regardless of data sensitivity). Second, modeling risk as the probabilistic union of base risk and uncertainty avoids arbitrary clipping and bounds $R(C)$ in $[0, 1]$. Uncertainty acts as an enhancer: incomplete telemetry ($U(C) > 0$) increases the consequence score even when base risk is low.

\subsection{Risk-Adaptive Quorum Predicate}
The quorum predicate $\Pi(C, Q, \mathit{seq}) = \text{True}$ iff the following six conditions are simultaneously satisfied:

\begin{enumerate}
    \item \textbf{Quorum Selection Integrity:} The selected quorum $Q$ must match the deterministic quorum selection function $H(C, S_{\mathit{seq}}, \mathit{registry\_snapshot})$:
    \begin{equation}
        Q = H(C, S_{\mathit{seq}}, \mathit{registry\_snapshot})
    \end{equation}
    where $\mathit{registry\_snapshot}$ is the immutable validator-registry snapshot active for epoch $\mathit{seq}$. The function $H$ is a deterministic constrained selection algorithm evaluated under $\mathsf{pred\_cfg}$. It filters and ranks validators by eligibility, domain authority weight $w_i(D)$, diversity contribution, veto-archetype coverage for $\mathcal{K}_{\text{crit}}(C)$, and a seed-derived tie-breaker $t_i=\textsf{PRF}_{h(C \parallel \mathit{seq})}(\mathit{validator\_id}_i)$. Starting from an empty prefix, $H$ greedily appends the highest-ranked eligible validator whose addition preserves the risk-adaptive correlation threshold $\epsilon(R(C))$ and diversity bounds $\beta_M,\beta_\kappa$, stopping only when the target quorum size and required veto-archetype coverage are satisfied. The seed is used only for deterministic tie-breaking; because all inputs are $C$, $S_{\mathit{seq}}$, $\mathit{registry\_snapshot}$, and the recorded predicate configuration, the execution gate can re-run $H$ exactly.
    
    \item \textbf{Diversity Condition:} The cognitive failure correlation between any two validators in the quorum is strictly bounded by a risk-adaptive correlation threshold $\epsilon(R(C))$:
    \begin{equation}
        \max_{i, j \in Q, i \neq j} \rho_{ij}(D) \le \epsilon(R(C))
    \end{equation}
    where $D$ represents the contract domain.
    
    \item \textbf{Weighted Approval Threshold:} The sum of weighted approvals exceeds a risk-adaptive confidence threshold $\tau(R(C))$:
    \begin{equation}
        \sum_{a_i \in Q} \mathbb{I}[v_i = +1] \cdot w_i \cdot c_i \ge \tau(R(C))
    \end{equation}
    where $\mathbb{I}[\cdot]$ is the indicator function.
    
    \item \textbf{Critical Archetype Veto:} No validator $a_i \in Q$ belonging to the critical veto-enabled archetype set $\mathcal{K}_{\text{crit}}(C)$ votes for rejection with high calibrated confidence:
    \begin{equation}
        \nexists a_i \in Q \text{ s.t. } \kappa_i \in \mathcal{K}_{\text{crit}}(C) \land v_i = -1 \land c_i \ge \theta_{\text{veto}}
    \end{equation}
    where $\theta_{\text{veto}}$ is the veto confidence threshold.
    
    \item \textbf{Signature Validity:} The cryptographic signatures are verified against the validators' public keys:
    \begin{equation}
        \forall a_i \in Q, \quad \text{Verify}(P_i, \textsf{VoteRecord}_i, \pi_i) = \text{True}
    \end{equation}
    
    \item \textbf{Evidence \& Sequence Epoch Binding:} All signatures bind to the same contract hash $h(C)$, evidence digest $h(E)$, and current sequence epoch $\mathit{seq}$:
    \begin{equation}
        \forall a_i \in Q, \quad \textsf{VoteRecord}_i[4{:}6] = (h(C), h(E), \mathit{seq})
    \end{equation}
    The identity and archetype fields in each $\textsf{VoteRecord}_i$ must also match the corresponding entry in $\mathit{registry\_snapshot}$.
\end{enumerate}

To protect high-risk operations, the weighted approval threshold $\tau(R(C))$ increases monotonically with $R(C)$, while the correlation limit $\epsilon(R(C))$ decreases. High-consequence contracts thus require higher validation consensus and lower quorum correlation.

\subsection{Quorum Proof}
Upon successful evaluation, the control plane compiles the signatures, votes, and sequence numbers into an auditable quorum proof $\Gamma$:
\begin{equation}
    \begin{aligned}
    \Gamma = (&Q, \{\textsf{VoteRecord}_i\}_{a_i \in Q}, \sigma_{\text{agg}},\\
    &h(C), h(E), \mathit{seq}, \mathsf{pred\_cfg})
    \end{aligned}
\end{equation}
Here, $Q$ is the ordered set of selected validator identities, $\{\textsf{VoteRecord}_i\}_{a_i \in Q}$ are the exact signed vote records, $\sigma_{\text{agg}}$ is the aggregate signature, $h(C)$ and $h(E)$ bind the contract and evidence, $\mathit{seq}$ binds the proof to the sequence epoch, and $\mathsf{pred\_cfg}$ identifies the predicate version, thresholds, diversity bounds, and veto configuration used for admission. This proof records cryptographically verifiable evidence that the quorum predicate was satisfied for sequence epoch $\mathit{seq}$ at admission time.

\subsection{Sovereign Execution Gate}
The sovereign execution gate decouples semantic validation from execution authority. Proposers and validators lack write access to the infrastructure state $S_{\mathit{seq}}$. Only the control plane $G$ executes admitted contracts, via the state transition function:
\begin{equation}
    S_{\mathit{seq}+1} = \begin{cases} 
      S_{\mathit{seq}} + \Delta S_C &
      \begin{aligned}[t]
      \text{if }& \text{Verify}(\Gamma, C, S_{\mathit{seq}},\\
                & \mathit{registry\_snapshot}) \\
                & \land\ \text{epoch}(\Gamma) = \mathit{seq} \\
                & \land\ \Phi_{\text{gate}}(C, \Gamma, S_{\mathit{seq}}) = 1
      \end{aligned} \\
      S_{\mathit{seq}}              & \text{otherwise}
    \end{cases}
\end{equation}
where $\Phi_{\text{gate}}$ is a boolean control representing the sovereign execution gate's policy and invariant verification, detailed in Section 8.2, and $\text{Verify}(\Gamma, C, S_{\mathit{seq}}, \mathit{registry\_snapshot})$ verifies:
\begin{enumerate}
    \item The aggregate signature is cryptographically valid under the registered public keys for the validators in $Q$ over the distinct signed messages $\{\textsf{VoteRecord}_i\}_{a_i \in Q}$.
    \item Each vote record binds to $h(C)$, $h(E)$, $\mathit{seq}$, $\mathit{validator\_id}_i$, and $\mathit{archetype}_i$ as registered in $\mathit{registry\_snapshot}$.
    \item The quorum identities in $\Gamma$ equal the exact subset returned by re-running $Q = H(C, S_{\mathit{seq}}, \mathit{registry\_snapshot})$ at the execution gate.
    \item The quorum predicate $\Pi(C, Q, \mathit{seq})$ is satisfied under $\mathsf{pred\_cfg}$.
\end{enumerate}
Re-evaluating $H$ directly at the execution gate forces a Byzantine control-plane aggregator to present signatures from the selected quorum rather than from an unauthorized or homogeneous subset of validators. Checking that $\text{epoch}(\Gamma) = \mathit{seq}$ prevents cross-epoch replay attacks, where an attacker attempts to execute an old contract approved under a previous system state. Cloud-native implementations can enforce this gate using mechanisms such as IAM session policies, service control policies (SCPs), CalledVia conditions, capability tokens, or Kubernetes webhook controllers.

\subsection{Formal Properties}
SQA separates a cryptographic admission property from a semantic assurance claim. The former is enforced by the sovereign gate and can be checked mechanically. The latter depends on calibrated validator behavior and empirical estimates of correlated cognitive failure.

\begin{quote}
\textbf{Property 1 (Gate Integrity).} \textit{If the sovereign execution gate is correctly enforced, no infrastructure mutation can be committed unless the request carries a valid execution contract $C$, evidence digest $h(E)$, active sequence epoch $\mathit{seq}$, selected validator quorum $Q$, valid signed vote records $\{\textsf{VoteRecord}_i\}_{a_i \in Q}$, a valid quorum proof $\Gamma$, and an authorized control-plane identity.}
\end{quote}

\begin{quote}
\textbf{Property 2 (Probabilistic Semantic Assurance).} \textit{Under calibrated validator failure probabilities and empirically estimated pairwise or joint failure correlations for domain $D$, the residual probability that SQA admits an unsafe contract is bounded by a risk function}
\begin{equation}
    \begin{aligned}
    &\Pr[\textsf{admit}(C) \mid C \text{ unsafe}]\\
    &\quad \le
    \textsf{ResidualRisk}(C,Q,\rho,c,w).
    \end{aligned}
\end{equation}
\textit{This risk function is estimated from held-out validation tasks or conservatively upper-bounded from the calibrated joint failure model. In the prototype reported here, it is estimated empirically by bootstrap over held-out validation tasks.}
\end{quote}

\noindent
For a selected quorum $Q$, let $X_i(C)=1$ denote the event that validator $a_i$ incorrectly approves an unsafe contract $C$ with vote record confidence $c_i$, and let $\mathcal{A}_Q(C,c,w)$ be the set of validator failure patterns that satisfy the weighted approval threshold and avoid any critical-archetype veto under $\mathsf{pred\_cfg}$. If a calibrated joint distribution $\widehat{P}_D$ over quorum failure patterns is available for domain $D$, we define:
\begin{equation}
    \begin{aligned}
    &\textsf{ResidualRisk}(C,Q,\rho,c,w) \\
    &\quad =
    \sum_{x \in \mathcal{A}_Q(C,c,w)}
    \widehat{P}_D(X_Q=x \mid C \text{ unsafe}).
    \end{aligned}
\end{equation}
When only marginal failure rates and pairwise correlations are available, a deployment may instead use a conservative upper envelope over distributions matching those estimates:
\begin{equation}
    \begin{aligned}
    &\textsf{ResidualRisk}(C,Q,\rho,c,w) \\
    &\quad =
    \sup_{\widehat{P} \in \mathcal{P}(\hat{p},\hat{\rho})}
    \sum_{x \in \mathcal{A}_Q(C,c,w)}
    \widehat{P}(X_Q=x \mid C \text{ unsafe}),
    \end{aligned}
\end{equation}
where $\mathcal{P}(\hat{p},\hat{\rho})$ is the family of joint distributions $\widehat{P}$ consistent with calibrated validator failure rates $\hat{p}$ and estimated correlations $\hat{\rho}$. In the prototype evaluation, this quantity is estimated empirically by bootstrap resampling over held-out validation tasks, preserving observed joint failure structure within each domain $D$.
Solving the upper-envelope optimization over all distributions consistent with marginal and pairwise estimates is a deployment option and a future-work direction for sharper conservative certification; it is not required for the prototype results in Section~\ref{sec:evaluation}.

\subsection{Proof Sketch}
We first prove Gate Integrity, then state the scope of semantic assurance.

\paragraph{Gate Integrity.}
Assume the sovereign execution gate mediates all mutative infrastructure operations, the control-plane signing keys and registry snapshot are not compromised, and the gate implementation rejects any request for which $\text{Verify}(\Gamma, C, S_{\mathit{seq}}, \mathit{registry\_snapshot})$ is false. Under these assumptions:
\begin{enumerate}
    \item \textbf{Validator Isolation:} Validator containers are executed in gVisor sandboxes with read-only overlay mounts and network interfaces disabled. Thus, validators cannot bypass the control plane to directly mutate $S_{\mathit{seq}}$.
    \item \textbf{Cryptographic Binding:} Each validator's signature $\pi_i$ binds the exact $\textsf{VoteRecord}_i$, including vote, assurance score, rationale reference, contract hash $h(C)$, evidence hash $h(E)$, sequence epoch $\mathit{seq}$, validator identity, and archetype. An adversary cannot reuse signatures for a different contract $C'$ or a different evidence trace $E'$.
    \item \textbf{Replay Prevention:} The sovereign gate verifies that the sequence epoch $\mathit{seq}$ bound within the validator signatures matches the current epoch of the system. An attacker attempting to replay a previously valid quorum proof $\Gamma$ is rejected after the active system epoch increments ($seq' > \mathit{seq}$), causing $\text{epoch}(\Gamma) = \mathit{seq}'$ to evaluate to false.
    \item \textbf{Signature Integrity:} The sovereign gate verifies the aggregate signature. Any modification to the contract, evidence, or votes invalidates the signature, causing the verification to fail.
    \item \textbf{Byzantine Aggregator Resistance:} Because the sovereign gate independently re-computes $Q = H(C, S_{\mathit{seq}}, \mathit{registry\_snapshot})$ using the state $S_{\mathit{seq}}$ and the active registry snapshot, any attempt by a compromised aggregator to present a proof signed by a non-selected or homogeneous set of validators is rejected during verification.
    \item \textbf{Gate Enforcement:} The cloud-native environment blocks all direct mutative actions unless they originate from the designated control-plane service role authenticated by the SQA proof $\Gamma$.
\end{enumerate}
Therefore, agent reasoning alone cannot mutate $S_{\mathit{seq}}$. Any committed state mutation must pass through the gate with a valid contract, evidence digest, active epoch, exact selected quorum, signed vote records, quorum proof, and authorized control-plane identity. This proves Property~1.

\paragraph{Semantic Assurance.}
Gate Integrity does not imply that every admitted contract is semantically safe. Validators are non-deterministic evaluators whose failures may be correlated by model family, retrieval context, prompt template, or shared training data. SQA therefore does not claim a mathematical guarantee of semantic safety. Instead, SQA cryptographically enforces a calibrated, auditable, risk-adaptive semantic certification process before state mutation. The residual probability of admitting an unsafe contract is modeled by $\textsf{ResidualRisk}(C,Q,\rho,c,w)$ and estimated from calibration and held-out validation data as described in Property~2. The empirical evaluation measures this residual risk under the studied workloads; deployments must continually recalibrate validator failure rates, confidence scores, weights, and correlation estimates as models and operating domains drift.

%% file: sections/07-protocol-design.tex
\section{Protocol Design}
\label{sec:protocol}

The SQA protocol governs the lifecycle of an infrastructure mutation, from an autonomous agent's proposal to final admission at the sovereign gate. This section gives the reference architecture, the admission phases, the quorum-selection procedure, and the signature aggregation mechanism.

\subsection{Protocol Phases}
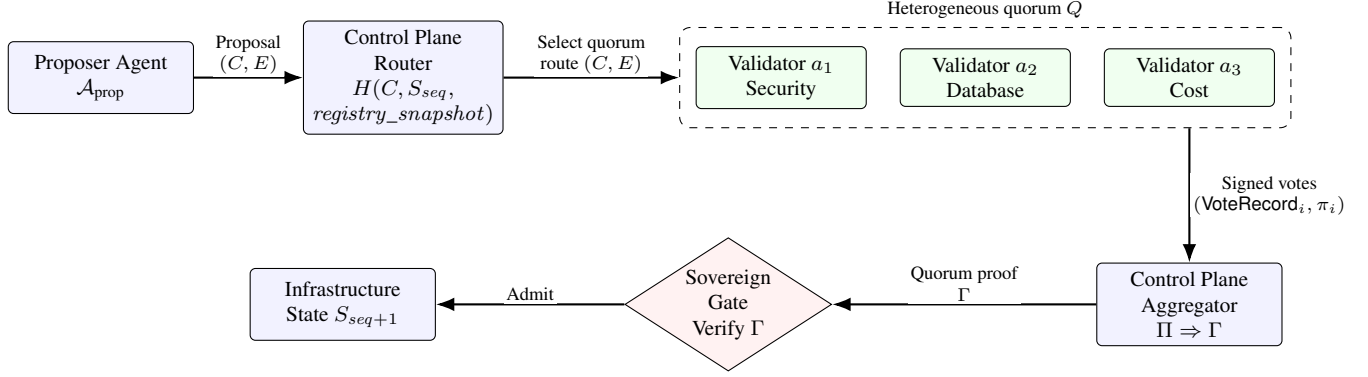
\begin{figure*}[htpb]
  \centering
  \begin{tikzpicture}[
    node distance=1.0cm and 1.0cm,
    component/.style={draw, fill=blue!5, rounded corners=2pt, minimum height=0.95cm, minimum width=2.45cm, align=center, font=\footnotesize},
    validator/.style={draw, fill=green!6, rounded corners=2pt, minimum height=0.68cm, minimum width=2.25cm, align=center, font=\footnotesize},
    gate/.style={draw, fill=red!5, diamond, aspect=1.55, inner sep=1pt, minimum width=1.95cm, align=center, font=\footnotesize},
    group/.style={draw, dashed, rounded corners=4pt, inner xsep=6pt, inner ysep=7pt},
    line/.style={draw, -{Latex[scale=1.05]}, thick},
    edge label/.style={font=\scriptsize, align=center, fill=white, inner sep=1pt}
  ]
    \node [component] (prop) at (0,1.45) {Proposer Agent\\$\mathcal{A}_{\text{prop}}$};
    \node [component] (router) at (4.0,1.45) {Control Plane\\Router\\$H(C,S_{\mathit{seq}},$\\$\mathit{registry\_snapshot})$};

    % Shifted the quorum group right to keep the routing arrow longer
    \node [validator] (val1) at (9.0,1.45) {Validator $a_1$\\Security};
    \node [validator] (val2) at (11.7,1.45) {Validator $a_2$\\Database};
    \node [validator] (val3) at (14.4,1.45) {Validator $a_3$\\Cost};
    \node [group, fit=(val1)(val2)(val3), label={[font=\scriptsize]above:Heterogeneous quorum $Q$}] (quorum) {};

    % Shifted aggregator right to align with the new val3 position
    \node [component] (agg) at (14.4,-1.55) {Control Plane\\Aggregator\\$\Pi \Rightarrow \Gamma$};
    \node [gate] (gate) at (8.3,-1.55) {Sovereign\\Gate\\Verify $\Gamma$};
    \node [component] (state) at (3.2,-1.55) {Infrastructure\\State $S_{\mathit{seq}+1}$};

    \draw [line] (prop) -- node[above, edge label] {Proposal\\$(C,E)$} (router);
    
    % Label moved above the line as requested
    \draw [line] (router) -- node[above, edge label] {Select quorum\\route $(C,E)$} (quorum.west);
    
    % Arrow starts from the outer box boundary directly below val3
    \draw [line] (quorum.south -| val3.south) -- node[right, edge label] {Signed votes\\$(\textsf{VoteRecord}_i,\pi_i)$} (agg.north);
    
    \draw [line] (agg) -- node[above, edge label] {Quorum proof\\$\Gamma$} (gate);
    \draw [line] (gate) -- node[above, edge label] {Admit} (state);
  \end{tikzpicture}
  \caption{Semantic Quorum Assurance reference architecture. The control plane routes a contract $C$ and evidence chain $E$ to a diversity-bounded validator quorum $Q$. Validators return signed vote records with calibrated assurance scores and rationale references; the aggregator emits quorum proof $\Gamma$ only when the risk-adaptive predicate is satisfied. The sovereign gate verifies $\Gamma$ before authorizing the state transition to $S_{\mathit{seq}+1}$.}
  \label{fig:sqa-architecture}
\end{figure*}

\begin{figure}[t]
  \centering
  \begin{tikzpicture}[
    node distance=0.55cm,
    stepnode/.style={
      draw=slate!40!black!30, 
      fill=blue!4!gray!5, 
      rectangle, 
      text width=0.88\columnwidth, 
      minimum height=0.65cm, 
      align=left, 
      font=\scriptsize, 
      rounded corners=3pt, 
      inner sep=5pt
    },
    arrow/.style={
      draw=gray!70!black, 
      -{Stealth[scale=1.1]}, 
      line width=1.1pt
    }
  ]
    % Nodes with alternating/progressive clean tints
    \node [stepnode, fill=blue!6!white!93] (s1) {1. \textbf{Propose}: Submit contract $C$ and evidence $E$};
    
    % Step 2 with the equation forced onto a new centered line
    \node [stepnode, fill=blue!4!white!95, below=of s1] (s2) {2. \textbf{Route}: Select diversity-bounded quorum:\\ \centerline{$Q=H(C,S_{\mathit{seq}},\mathit{registry\_snapshot})$}};
    
    \node [stepnode, fill=teal!3!white!96, below=of s2] (s3) {3. \textbf{Evaluate}: Run validators in read-only sandboxes};
    \node [stepnode, fill=teal!4!white!95, below=of s3] (s4) {4. \textbf{Vote}: Return $\textsf{VoteRecord}_i$ signed as $\pi_i$};
    \node [stepnode, fill=emerald!3!white!96, below=of s4] (s5) {5. \textbf{Aggregate}: Verify $\Pi(C,Q,\mathit{seq})$ under $\mathsf{pred\_cfg}$ and construct $\Gamma$};
    \node [stepnode, fill=emerald!5!white!94, below=of s5] (s6) {6. \textbf{Gate}: Verify $\Gamma$ and authorize $\Delta S_C$};

    % Connecting paths
    \path [arrow] (s1) -- (s2);
    \path [arrow] (s2) -- (s3);
    \path [arrow] (s3) -- (s4);
    \path [arrow] (s4) -- (s5);
    \path [arrow] (s5) -- (s6);
  \end{tikzpicture}
  \caption{SQA protocol lifecycle. Each proposal passes through routing, sandboxed evaluation, signed voting, aggregation, and sovereign gate verification before execution.}
  \label{fig:protocol-lifecycle}
\end{figure}
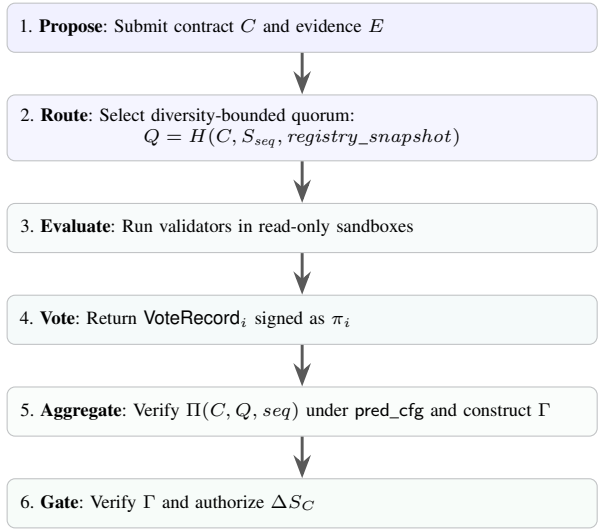

Figure~\ref{fig:sqa-architecture} shows the control-plane boundary that separates proposal, validation, aggregation, and execution. Figure~\ref{fig:protocol-lifecycle} then summarizes the six sequential admission phases.

\begin{enumerate}
    \item \textbf{Proposal Phase:} The proposer agent $\mathcal{A}_{\text{prop}}$ identifies an operational need and compiles an execution contract $C = (\Delta S_C, I_C)$ alongside the cryptographic evidence chain $E$. It submits this bundle to the control plane.
    
    \item \textbf{Dissemination \& Routing Phase:} The control plane parses the execution contract, determines the consequence score $R(C)$, and determines the required risk-adaptive threshold $\tau(R(C))$. Using the quorum selection algorithm (Section~\ref{subsec:quorum-selection}), the control plane dynamically selects a quorum $Q$ of validators that satisfy the diversity constraint $\mathcal{D}(Q)$. It then distributes $C$ and $E$ to each validator $a_i \in Q$.
    
    \item \textbf{Evaluation Phase:} Each validator $a_i$ is launched in an isolated, read-only container sandbox. The validator receives the contract $C$, the evidence chain $E$, and its authorized state projection $S_{\mathit{seq}} \restriction_{R_i}$. It executes $\sigma_i(C, E, S_{\mathit{seq}} \restriction_{R_i})$ to determine whether the proposed mutation is consistent with the relevant safety, compliance, and operational invariants.
    
    \item \textbf{Voting Phase:} The validator generates the canonical vote record and signs exactly that object:
    \begin{equation}
        \begin{aligned}
        \textsf{VoteRecord}_i = (&v_i, c_i, r_i, h(C), h(E), \mathit{seq},\\
        &\mathit{validator\_id}_i, \mathit{archetype}_i)
        \end{aligned}
    \end{equation}
    \begin{equation}
        \pi_i = \text{Sign}_{K_i}(\textsf{VoteRecord}_i)
    \end{equation}
    The validator returns this signed vote record to the control plane aggregator and immediately terminates.
    
    \item \textbf{Aggregation Phase:} The control plane gathers the signed vote records and verifies them against the registered validator public keys. It then evaluates the SQA quorum predicate $\Pi(C,Q,\mathit{seq})$ under $\mathsf{pred\_cfg}$. If the predicate is satisfied, the control plane constructs the aggregated assurance proof $\Gamma$.
    
    \item \textbf{Submission \& Gatekeeping Phase:} The control plane submits the contract $C$ and the proof $\Gamma$ to the sovereign execution gate. The gate cryptographically verifies $\Gamma$, checks that the invariants hold, and triggers the physical state change:
    \begin{equation}
        S_{\mathit{seq}+1} = S_{\mathit{seq}} + \Delta S_C
    \end{equation}
\end{enumerate}

\subsection{Quorum Selection Algorithm}
\label{subsec:quorum-selection}
To prevent correlated cognitive failures, the control plane runs the deterministic constrained selection algorithm $H(C,S_{\mathit{seq}},\mathit{registry\_snapshot})$ to build the validator quorum $Q$. For a fixed predicate configuration $\mathsf{pred\_cfg}$, the algorithm uses the registry snapshot active at epoch $\mathit{seq}$, the target contract $C$, diversity bounds $\beta_M,\beta_\kappa$, and the deterministic seed $s=h(C \parallel \mathit{seq})$. The seed is used only as a tie-breaker, so the sovereign execution gate can re-run the same algorithm exactly.
\begin{enumerate}
    \item Derive the target domain $D$ of the contract, the consequence score $R(C)$, the target quorum size $N_Q$, and the required veto-archetype set $\mathcal{K}_{\text{crit}}(C)$.
    \item Filter registry entries by eligibility: active public key, non-revoked status, authorized read set for the contract domain, available sandbox backend, and compatibility with the current predicate configuration.
    \item Starting from an empty quorum prefix, rank each remaining eligible validator by the lexicographic key consisting of domain authority weight $w_i(D)$, marginal diversity contribution with respect to the current prefix, uncovered veto-archetype coverage, and tie-breaker $\textsf{PRF}_{s}(\mathit{validator\_id}_i)$.
    \item Greedily append the highest-ranked candidate whose addition preserves:
    \begin{itemize}
        \item the pairwise correlation threshold $\max_{i,j \in Q, i\neq j}\rho_{ij}(D) \le \epsilon(R(C))$;
        \item the model-family and archetype diversity bounds $\beta_M,\beta_\kappa$;
        \item progress toward covering all required critical veto archetypes.
    \end{itemize}
    \item Stop when $|Q|=N_Q$ and the required veto-archetype coverage is satisfied. If no feasible quorum exists, $H$ returns failure and the contract is rejected or escalated rather than routed to an ad hoc quorum.
\end{enumerate}

\subsection{Cryptographic Signature Aggregation}
To minimize proof size at the execution gate, SQA leverages aggregate signatures (e.g., BLS signatures~\cite{boneh-multisig}). Instead of carrying $N_Q$ independent signature objects through the admission path, the validator signatures are aggregated into a single signature $\sigma_{\text{agg}}$:
\begin{equation}
    \sigma_{\text{agg}} = \textsf{BLS.Aggregate}(\{\pi_i\}_{a_i \in Q})
\end{equation}
The sovereign execution gate verifies this aggregate signature against the selected validator public keys and the signed vote records:
\begin{equation}
    \begin{aligned}
    &\textsf{BLS.AggregateVerify}(\{P_i\}_{a_i \in Q},\\
    &\quad \{\textsf{VoteRecord}_i\}_{a_i \in Q}, \sigma_{\text{agg}}) = \text{True}
    \end{aligned}
\end{equation}
BLS aggregate verification checks a list of distinct signed messages against the corresponding registered public keys. To prevent rogue-key attacks, validator public keys require proof-of-possession at registration time. The gate verifies aggregates only against public keys read from the immutable $\mathit{registry\_snapshot}$ for the selected quorum $Q$; keys carried in a proof are ignored. The aggregation is an elliptic-curve group operation over signatures, not ordinary integer multiplication, keeping the gate-facing proof compact while preserving per-validator accountability.

%% file: sections/08-sovereign-execution-gating.tex
\section{Sovereign Execution Gating}
\label{sec:gating}

SQA strictly decouples \textbf{semantic approval} from \textbf{execution authority}. Traditional agentic systems grant agents direct credentials (e.g., cloud access keys or Kubernetes service account tokens) to evaluate and apply changes. This design violates least privilege: a compromised or errant agent can execute destructive actions. SQA routes all mutations through a sovereign execution gate, $\Phi_{\text{gate}}$, ensuring validators evaluate proposals but possess \textbf{no execution authority}.

\subsection{Decoupled Validation and Execution}
Decoupling validation from execution uses sandboxed validator isolation (Figure~\ref{fig:sqa-architecture}). Quorum validators $a_i \in Q$ run in ephemeral sandboxes configured with:
\begin{itemize}
    \item \textbf{Read-Only Filesystem with Pre-Fetched State:} The validator sandbox has no dynamic network access to target systems. Instead, to evaluate the state projection $S_{\mathit{seq}} \restriction_{R_i}$, the trusted control plane pre-fetches the required system context (e.g., current resource configurations, active metrics, schema logs) matching the validator's read set $R_i$, serializes it as a JSON payload, and mounts it as a read-only data volume inside the sandbox.
    \item \textbf{Disabled Outbound Network:} All standard network interfaces are disabled inside the container (\texttt{--net=none}). Validators cannot dynamically query external databases, call cloud provider endpoints, or communicate with public servers. They can only issue model inference calls through a local secure UNIX domain socket connected to a model proxy managed by the control plane, and output their final signed vote record $(\textsf{VoteRecord}_i, \pi_i)$ via secure IPC.
    \item \textbf{Ephemeral Lifetime:} The container hosting the validator is destroyed immediately after the vote is cast, wiping all pre-fetched state and runtime memory.
\end{itemize}

Lacking write credentials and outbound network interfaces, validators cannot execute mutations. They remain analysis-only components.

\subsection{Sovereign Execution Gate}
Mutative authority resides solely within the execution gate, $\Phi_{\text{gate}}$, a trusted component running in an isolated environment (such as a hypervisor partition or TEE). 

The gate exposes a verification function, $\text{Verify}(\Gamma, C, S_{\mathit{seq}}, \mathit{registry\_snapshot})$, which recomputes the expected quorum $Q$, checks validator public keys, verifies the BLS aggregate signature over the signed vote records, and confirms that the contract hash, evidence digest, active sequence epoch, validator identities, archetypes, and predicate configuration match the proof. The state transition logic is:
\begin{equation}
    S_{\mathit{seq}+1} = 
    \begin{cases} 
      S_{\mathit{seq}} + \Delta S_C &
      \begin{aligned}[t]
      \text{if }& \text{Verify}(\Gamma, C, S_{\mathit{seq}},\\
                & \mathit{registry\_snapshot}) = \text{True}\\
                & \land\ \text{epoch}(\Gamma) = \mathit{seq}\\
                & \land\ \Phi_{\text{gate}}(C,\Gamma,S_{\mathit{seq}}) = 1
      \end{aligned} \\
      S_{\mathit{seq}} & \text{otherwise}
    \end{cases}
\end{equation}
where $\Phi_{\text{gate}}(C,\Gamma,S_{\mathit{seq}}) \in \{0, 1\}$ is a Boolean gate control. The gate control is set to $1$ only if:
\begin{enumerate}
    \item The contract's declarative invariants $I_C$ are verified to be structurally sound.
    \item The system is not under an administrative freeze or lock.
    \item The caller proposing the contract is authenticated and has execution permissions for the target resources.
\end{enumerate}

If the verification of the aggregated proof $\Gamma$ fails, or if the gate conditions are not satisfied, the state remains unchanged ($S_{\mathit{seq}+1} = S_{\mathit{seq}}$). The proposed mutation is rejected, and the control plane logs the validation failure for administrative review. This design prevents a compromised proposer or validator set from obtaining write authority unless the final execution gate verifies the exact selected quorum, the required cryptographic bindings, and the policy boundaries.

%% file: sections/09-implementation.tex
\section{Implementation}
\label{sec:implementation}

We implemented a prototype of the SQA control plane within the \openkedge framework. The implementation targets Kubernetes-managed cloud-native environments and is divided into three components: the sandboxed validation environment, the contract router/aggregator, and the Kubernetes admission controller gating mechanism.

\subsection{Sandboxed Validator Isolation}
To prevent malicious or compromised validator processes from interacting with the host system, validators run inside ephemeral sandboxes powered by \textbf{gVisor}~\cite{gvisor-docs} with a \texttt{runsc} runtime. gVisor intercepts system calls from the container and filters them through a guest kernel, reducing the host kernel surface exposed to validator code.

Each validator instance is configured as follows:
\begin{itemize}
    \item \textbf{Filesystem isolation:} The container root filesystem is mounted as read-only (\texttt{ro}) using an overlay mount. The pre-fetched state projection $S_{\mathit{seq}} \restriction_{R_i}$ is serialized as a JSON document and mounted at a secure, read-only path (e.g., \texttt{/etc/sqa/state.json}) inside the container. The validator's final output vote and signature are written to a shared memory partition (\texttt{tmpfs}) that is immediately wiped upon container termination.
    \item \textbf{Network isolation:} The container is spawned with the network stack disabled (\texttt{--net=none}), preventing validators from communicating with the public internet or external API endpoints. To fetch model reasoning, the validator communicates via an internal gRPC socket linked to a local model proxy server.
    \item \textbf{Model Proxy Security Filtering:} To prevent a compromised validator from exfiltrating sensitive telemetry (from $S_{seq} \restriction_{R_i}$) to an external LLM provider, the model proxy server acts as an egress filter. The proxy strips local environment credentials and metadata, then runs outbound prompts through regex-based and classifier-based anonymization filters that redact sensitive identifiers, API keys, internal IPs, and high-entropy strings before forwarding the request to the external LLM API.
\end{itemize}

\subsection{Execution Contracts \& Schema Definition}
Execution contracts are structured using YAML format. The schema enforces strict typing and declarative resource definition. Listing~\ref{lst:contract} illustrates an execution contract proposed by a database pruning agent:

\begin{lstlisting}[float,label=lst:contract,caption={Example of an SQA Execution Contract in YAML format.}]
apiVersion: openkedge.io/v1alpha1
kind: ExecutionContract
metadata:
  name: prune-backup-volumes
  uuid: "9f82d-4812a-431f-bc87-920f"
spec:
  target:
    resource: "persistentvolumeclaim"
    namespace: "database"
    selector:
      app: "mysql-backup"
  action:
    type: "PruneFiles"
    parameters:
      path: "/backups/historical/"
      threshold_days: 30
  invariants:
    - name: "active-restores-exist"
      query: "count(persistentvolumeclaims[app='mysql-backup']) > 0"
    - name: "system-read-only-false"
      query: "database.status == 'ONLINE'"
\end{lstlisting}

The evidence chain $E$ is compiled as a JSON-LD document containing input metrics, shell logs, and structured rationale artifacts leading to this proposal. It is bound to the contract via a SHA-256 hash digest.

\subsection{Router and Aggregate Signatures}
The SQA router and aggregator are written in Go. The router queries the validator registry $V$, records the active $\mathit{registry\_snapshot}$, parses the contract's consequence score, and selects the quorum.

For signature aggregation, we use the BLS12-381 pairing-friendly elliptic curve library. When a validator $a_i$ completes its evaluation, it constructs
\begin{equation}
    \begin{aligned}
    \textsf{VoteRecord}_i = (&v_i, c_i, r_i, h(C), h(E), \mathit{seq},\\
    &\mathit{validator\_id}_i, \mathit{archetype}_i)
    \end{aligned}
\end{equation}
and signs exactly $\textsf{VoteRecord}_i$ using its BLS private key. The Go aggregator aggregates the validator signatures with the BLS aggregate-signature API to produce $\sigma_{\text{agg}}$, then emits the quorum proof $\Gamma$ containing $Q$, the vote records, $\sigma_{\text{agg}}$, $h(C)$, $h(E)$, $\mathit{seq}$, and $\mathsf{pred\_cfg}$ before submission to the gate.

\subsection{Sovereign Gating with Kubernetes Admission Controllers}
The sovereign execution gate is integrated into the Kubernetes control plane as a Validating Admission Webhook.
When a proposer agent attempts to modify a resource, the request is intercepted by the Kubernetes API server and routed to the SQA admission gate:
\begin{enumerate}
    \item The gate extracts the \texttt{ExecutionContract} and the aggregate signature proof $\Gamma$ from the annotations of the request.
    \item It queries the validator registry snapshot to fetch the public keys of the quorum members.
    \item It re-runs $Q = H(C,S_{\mathit{seq}},\mathit{registry\_snapshot})$ and verifies that the quorum identities in $\Gamma$ match exactly.
    \item It verifies the aggregate signature over the distinct $\textsf{VoteRecord}_i$ messages using the BLS12-381 library and the registered public keys.
    \item It verifies that the SQA quorum predicate $\Pi(C,Q,\mathit{seq})$ is satisfied under $\mathsf{pred\_cfg}$.
\end{enumerate}
If verification succeeds, the admission controller approves the request, and the API server persists the state change to \texttt{etcd}. If it fails, the request is blocked and logged, returning an HTTP 403 Forbidden error to the calling proposer agent.

%% file: sections/10-evaluation.tex
\section{Evaluation}
\label{sec:evaluation}

We evaluate SQA on a controlled corpus of synthetic and incident-pattern-derived, infrastructure-inspired mutation scenarios. The evaluation measures whether the gate-facing assurance process reduces unsafe approvals, what latency and token cost it adds, and which quorum mechanisms contribute to the result. The headline safety numbers below are computed on held-out safe/unsafe trials, excluding ambiguous scenarios; calibration data are used only to set validator weights, confidence calibration, and empirical correlation estimates.

\subsection{Research Questions}
\label{subsec:research-questions}
We organize the evaluation around four research questions:
\begin{itemize}
    \item \textbf{RQ1:} Does SQA reduce unsafe approvals compared with single-agent validation, static policy, and homogeneous quorum baselines?
    \item \textbf{RQ2:} How much latency and cost does SQA add across low-, medium-, and high-risk contracts?
    \item \textbf{RQ3:} How much do model-family diversity, archetype diversity, vetoes, and calibrated confidence weighting each contribute?
    \item \textbf{RQ4:} How does SQA behave under increasing correlated validator failure?
\end{itemize}

\subsection{Dataset Construction}
\label{subsec:dataset-construction}
The dataset contains 500 infrastructure-inspired mutation scenarios spanning three operational domains: Kubernetes administration (180 scenarios), database administration (160 scenarios), and IAM/network security (160 scenarios). Each scenario contains an initial state snapshot $S_{\mathit{seq}}$, an execution contract $C=(\Delta S_C,I_C)$, an evidence chain summary $E$, and an expected semantic label. We label 310 scenarios as safe, 150 as unsafe, and 40 as ambiguous. Ambiguous cases are retained for sensitivity analysis but excluded from the main binary safety metrics.

Scenarios are generated from parameterized templates rather than live production traces. Template families cover Kubernetes pod scaling, namespace deletion, daemonset rollout, node drain, persistent-volume modification, SQL migration, backup pruning, index creation, user grant modification, IAM policy editing, security-group modification, and secret rotation. The unsafe templates encode context-dependent hazards such as deleting the only active backup, opening administrative ports to public CIDR ranges, granting wildcard IAM privileges, dropping a table before a verified snapshot, or draining all nodes that host a singleton control-plane dependency. Ninety-six unsafe scenarios are generalized from public incident patterns and cloud-operation anti-patterns; the concrete resource names, logs, and configurations are synthetic. Appendix~\ref{app:scenario-provenance} lists the scenario template families and public incident-pattern categories used to construct the unsafe scenarios.

Labels are assigned by deterministic outcome predicates over $(S_{\mathit{seq}},\Delta S_C,I_C,E)$ and then checked by two authors. Disagreements are resolved by inspecting the invariant that would be violated after execution. A scenario is labeled safe when the proposed mutation preserves all declared invariants under the supplied state and evidence. It is labeled unsafe when executing $\Delta S_C$ would violate an invariant, remove required recovery capacity, create an unintended privilege escalation, or expose sensitive data. Ambiguous scenarios intentionally omit enough telemetry that a cautious validator should reject or request escalation, but the ground truth cannot be reduced to a binary safe/unsafe label without additional observations.

\noindent\textbf{Scenario examples.}
\begin{center}
  \scriptsize
  \begin{tabular}{@{}p{0.16\columnwidth}p{0.43\columnwidth}p{0.25\columnwidth}@{}}
    \toprule
    \textbf{Label} & \textbf{Mutation} & \textbf{Expected behavior} \\
    \midrule
    Safe & Scale a stateless deployment from 4 to 6 replicas when pod disruption budgets, quotas, and health checks are satisfied. & Approve after verifying capacity and invariants. \\
    Unsafe & Prune files under a backup volume when $S_{\mathit{seq}}$ indicates that the same volume contains the only restorable snapshot for the primary database. & Reject due to loss of recovery capacity. \\
    Ambiguous & Rotate a database credential when the evidence chain includes the target secret but omits consumer rollout status. & Reject or escalate until missing telemetry is supplied. \\
    \bottomrule
  \end{tabular}
\end{center}

We use a stratified split with 100 calibration scenarios (60 safe, 30 unsafe, 10 ambiguous) and 400 held-out test scenarios (250 safe, 120 unsafe, 30 ambiguous). The calibration split sets validator reliability weights $w_i$, confidence calibration curves, consequence-score thresholds, and empirical correlation estimates. All safety tables report held-out safe/unsafe results, excluding ambiguous scenarios, unless stated otherwise. Because validators are stochastic, each held-out safe/unsafe scenario is evaluated under five fixed random seeds, producing 1,850 test trials for the primary metrics.

\subsection{Validator and Baseline Configuration}
\label{subsec:validator-baselines}
The SQA validator registry uses anonymized model classes rather than provider-specific model names. In our prototype evaluation, $M_A$ and $M_B$ are frontier general-purpose instruction models from different providers, $M_C$ is a compact low-latency instruction model, and $M_D$ is a locally hosted code-oriented model. Validators run with temperature $0.1$, top-$p=0.9$, a 20 second per-call timeout, and at most two retries on malformed JSON output. The retry prompt repeats the required schema but does not reveal the expected label.
We anonymize model classes in the paper because provider model identifiers and served versions can drift during review; the frozen artifact metadata records the exact identifiers separately without making the manuscript brittle or provider-promotional.

Validator archetypes are implemented as fixed system-prompt blocks. The prompt template is identical across validators except for the archetype block and read-set description. The archetypes used in the full SQA configuration are Security Auditor, Kubernetes/SRE Operator, Database Administrator, IAM/Compliance Reviewer, Cost/Reliability Reviewer, and Backup/Recovery Reviewer. Each prompt requires the validator to inspect $C$, $h(E)$, its authorized projection of $S_{\mathit{seq}}$, the relevant invariants $I_C$, and then emit $(v_i,c_i,r_i)$ in a structured JSON object that is serialized into $\textsf{VoteRecord}_i$.

We compare against the following baselines:
\begin{itemize}
    \item \textbf{Single-Agent Validation (SAV):} The proposer evaluates its own contract with a three-step self-reflection prompt: summarize intended effect, list possible invariant violations, then decide approve/reject. The model class and decoding settings match the proposer used to generate the contract.
    \item \textbf{Static Policy Engine (OPA):} We implement Rego rules for structural denials: privileged Kubernetes containers, namespace deletion without an allow-list annotation, persistent-volume deletion without a backup tag, IAM wildcard actions on wildcard resources, public ingress to administrative ports, database \texttt{DROP}/\texttt{TRUNCATE} without a declared backup reference, and secret updates without rotation metadata. OPA receives only the structured request and metadata, not the free-form evidence chain.
    \item \textbf{Homogeneous Quorum (HomoQ):} The quorum size and threshold schedule match SQA, but every validator uses model class $M_C$ with independent decoding seeds. This isolates quorum size from model-family diversity.
    \item \textbf{Human-in-the-Loop (HITL):} We do not conduct a human-subject review study. HITL is included only as a latency comparator with an assumed queued-review distribution fixed for the prototype evaluation: 12 minutes p50, 45 minutes p95, and 120 minutes p99. These values are modeling parameters for an interrupt-driven on-call approval workflow, not measurements from this experiment.
\end{itemize}

\subsection{Metrics}
\label{subsec:metrics}
We report unsafe approval rate, false rejection rate, precision, recall, and F1. For precision/recall/F1, the positive class is ``reject an unsafe contract.'' Thus, unsafe approval rate is the false-negative rate on unsafe scenarios, and false rejection rate is the false-positive rate on safe scenarios. We also report end-to-end validation latency from contract submission to construction of $\Gamma$, average token cost per contract, and quorum size distribution.

Latency is measured on a single prototype deployment with validators executed in parallel and sandbox images pre-warmed. Token cost is computed from recorded prompt and completion token counts using a fixed price sheet; the dollar values should be interpreted as reproducibility aids rather than stable deployment costs. The held-out consequence-score distribution selects quorum sizes as follows: 164 contracts use $|Q|=2$, 147 use $|Q|=4$, and 89 use $|Q|=7$.

All confidence intervals are 95\% stratified bootstrap intervals over held-out scenario IDs with 10,000 resamples. For each resampled scenario, all five stochastic trials are included to preserve within-scenario correlation.

\subsection{Reproducibility and Artifact Statement}
\label{subsec:reproducibility-artifacts}
The reported safety, latency, and cost results correspond to the prototype evaluation. The scenario split, validator archetype prompts, decoding parameters, retry policy, static OPA rules, consequence-score thresholds, confidence-calibration parameters, model-class assignments, and token-price sheet are fixed for all tables in this section. Stochastic validator calls are evaluated under five fixed random seeds per held-out scenario; bootstrap confidence intervals resample scenario identifiers, not individual vote records, to preserve within-scenario dependence.
The frozen artifact metadata records exact model identifiers, provider-reported model versions where available, prompt templates, decoding settings, retry policy, and the token-price sheet used for the prototype cost calculations.

For public timestamping and workshop review, we plan to release a companion artifact containing scenario templates, anonymized scenarios, labels, validator archetype prompts, static OPA/Rego baseline rules, ablation configurations, bootstrap scripts, artifact metadata with exact model identifiers, and aggregate result tables. We also plan to release the prototype router, aggregator, and Kubernetes admission-gate code as a reference implementation after removing deployment credentials and provider-specific configuration. We will not release API keys, raw provider telemetry, or organization-specific infrastructure logs.

\subsection{Results Tables}
\label{subsec:results-tables}
Table~\ref{tab:main-safety-results} answers RQ1. Full SQA reduces unsafe approval from 18.5\% for SAV to 0.3\% on the held-out unsafe trials, a 98.4\% relative reduction. Static OPA has low false rejection but misses context-dependent hazards that are not visible in structural request fields. HomoQ improves accountability over SAV but remains sensitive to correlated model-family failures.

\begin{table}[H]
  \centering
  \scriptsize
  \setlength{\tabcolsep}{2pt}
  \renewcommand{\arraystretch}{0.92}
  \resizebox{\columnwidth}{!}{%
  \begin{tabular}{lccccc}
    \toprule
    \textbf{Method} & \textbf{Unsafe} & \textbf{False Rej.} & \textbf{Prec.} & \textbf{Rec.} & \textbf{F1} \\
    \midrule
    Static OPA & 34.7 [30.5, 38.9] & 2.1 [1.3, 3.1] & 93.3 & 65.3 & 76.8 \\
    SAV & 18.5 [15.7, 21.6] & 8.7 [7.0, 10.6] & 81.8 & 81.5 & 81.6 \\
    HomoQ & 21.0 [17.8, 24.5] & 6.8 [5.3, 8.5] & 84.8 & 79.0 & 81.8 \\
    SQA & \textbf{0.3 [0.0, 0.8]} & 3.1 [2.1, 4.2] & 93.9 & \textbf{99.7} & \textbf{96.7} \\
    HITL & -- & -- & -- & -- & -- \\
    \bottomrule
  \end{tabular}}
  \caption{Main safety results on held-out safe/unsafe trials, excluding ambiguous scenarios, in the prototype. Brackets are 95\% bootstrap CIs; HITL was not measured for safety.}
  \label{tab:main-safety-results}
\end{table}

Table~\ref{tab:latency-cost} answers RQ2. The 1.45--4.12 second headline latency range corresponds to median SQA latency from low-risk to high-risk buckets. Tail latency grows with quorum size because $\Gamma$ construction waits for enough validator responses to satisfy $\Pi(C,Q,\mathit{seq})$.

\begin{table}[H]
  \centering
  \scriptsize
  \setlength{\tabcolsep}{2pt}
  \renewcommand{\arraystretch}{0.92}
  \resizebox{\columnwidth}{!}{%
  \begin{tabular}{lccccc}
    \toprule
    \textbf{Bucket} & \textbf{n} & \textbf{$|Q|$} & \textbf{p50 / p95 / p99} & \textbf{Cost} & \textbf{Tokens} \\
    \midrule
    Low & 164 & 2 & 1.45s / 2.08s / 2.64s & \$0.0048 & 8.1k \\
    Medium & 147 & 4 & 2.36s / 3.42s / 4.31s & \$0.026 & 32.4k \\
    High & 89 & 7 & 4.12s / 5.84s / 7.30s & \$0.080 & 82.7k \\
    HITL & -- & 1 reviewer & 720s / 2700s / 7200s & -- & -- \\
    \bottomrule
  \end{tabular}}
  \caption{Latency and token cost by consequence-score bucket in the prototype; HITL is an assumed queued-review comparator.}
  \label{tab:latency-cost}
\end{table}

Table~\ref{tab:ablation} answers RQ3. Each preliminary ablation changes one component of the full SQA configuration while keeping the held-out scenarios, random seeds, and consequence-score thresholds fixed. Removing model-family and archetype diversity has the largest effect on unsafe approvals. Removing vetoes mainly affects high-consequence contracts, while disabling confidence weighting increases both unsafe approvals and false rejections because all validator outputs are treated as equally reliable.

\begin{table}[H]
  \centering
  \scriptsize
  \setlength{\tabcolsep}{3pt}
  \renewcommand{\arraystretch}{0.92}
  \begin{tabular}{lccc}
    \toprule
    \textbf{Configuration} & \textbf{Unsafe} & \textbf{False Rej.} & \textbf{F1} \\
    \midrule
    Full SQA & \textbf{0.3} & 3.1 & \textbf{96.7} \\
    No diversity & 6.9 & 4.0 & 92.9 \\
    No veto & 3.7 & \textbf{2.6} & 95.8 \\
    No confidence weighting & 2.1 & 5.8 & 94.0 \\
    Homogeneous model family & 21.0 & 6.8 & 81.8 \\
    Generic archetypes & 14.8 & 15.1 & 78.8 \\
    \bottomrule
  \end{tabular}
  \caption{Preliminary ablations on held-out safe/unsafe trials, excluding ambiguous scenarios, in the prototype. The positive class for F1 is unsafe-contract rejection.}
  \label{tab:ablation}
\end{table}

\subsection{Correlated Failure Stress Test}
\label{subsec:correlated-failure-stress}
To answer RQ4, we estimate empirical pairwise failure correlation on the calibration split using the binary failure indicators $F_i(C)$ from Section~\ref{subsec:correlated-failure}. For validators $a_i$ and $a_j$ in domain $D$, $\hat{\rho}_{ij}(D)$ is the sample correlation of their centered failure indicators over calibration scenarios. We then stress-test admission by resampling held-out validator outcomes with a Gaussian-copula approximation whose target pairwise correlation is swept over $\rho \in \{0,0.25,0.5,0.75,1.0\}$. This stress test is not a new semantic benchmark; it isolates the effect of correlated failures on the quorum decision rule, with the results summarized in Table~\ref{tab:correlated-stress}.

\begin{table}[H]
  \centering
  \scriptsize
  \setlength{\tabcolsep}{3pt}
  \renewcommand{\arraystretch}{0.92}
  \begin{tabular}{cccc}
    \toprule
    \textbf{$\rho$} & \textbf{SQA Unsafe} & \textbf{HomoQ Unsafe} & \textbf{SQA False Rej.} \\
    \midrule
    0.00 & 0.1 & 4.9 & 3.0 \\
    0.25 & 0.4 & 8.7 & 3.2 \\
    0.50 & 0.8 & 13.6 & 3.4 \\
    0.75 & 2.6 & 18.8 & 3.8 \\
    1.00 & 8.4 & 21.0 & 4.1 \\
    \bottomrule
  \end{tabular}
  \caption{Correlated failure stress test in the prototype. Full correlation substantially degrades SQA.}
  \label{tab:correlated-stress}
\end{table}

The stress test confirms the expected boundary: SQA reduces residual risk when validator failures are not perfectly coupled, but it does not guarantee semantic safety under fully correlated cognitive failure. This is consistent with Property~2; the empirical residual risk must be re-estimated when model families, prompts, retrieval corpora, or operating domains change.

\subsection{Threats to Validity}
\label{subsec:evaluation-threats}
\textbf{Synthetic dataset limitations.} The scenarios are generated from templates and generalized incident patterns, not from a live production change stream. They may underrepresent rare interactions among cloud services, organizational policies, and deployment-specific conventions.

\textbf{Labeler bias.} Labels depend on author-defined invariants and adjudication rules. We reduce ambiguity by using deterministic outcome predicates, but different organizations may classify some mutations differently depending on risk tolerance and recovery procedures.

\textbf{Model-version drift.} The anonymized model classes may change behavior as providers update serving stacks. Calibration parameters, confidence mappings, and $\rho$ estimates should therefore be treated as time-bounded measurements rather than stable constants.

\textbf{Prompt sensitivity.} Validator archetype prompts are fixed in this evaluation. Alternate prompt wording, longer evidence windows, or different output schemas could change both unsafe approval and false rejection rates.

\textbf{Pricing drift.} Token costs are computed from a frozen price sheet and recorded token counts. Provider pricing, caching discounts, batching, and local inference can change the dollar values without changing the protocol.

\textbf{Limited domain coverage.} The dataset covers Kubernetes, database administration, and IAM/network security. It does not evaluate data-plane operations, ML training pipelines, financial transaction systems, or safety-critical cyber-physical infrastructure.

%% file: sections/11-discussion.tex
\section{Discussion}
\label{sec:discussion}

The implementation and evaluation of SQA highlight several engineering trade-offs, challenges, and future research directions for securing non-deterministic infrastructure control planes.

\subsection{The Challenge of Score Calibration}
The effectiveness of the SQA quorum predicate depends heavily on the accuracy of the calibrated assurance scores $c_i$. If validators output poorly calibrated confidence metrics (e.g., claiming 99\% confidence when their semantic judgment is incorrect), they can distort the weighted voting process. 

While our calibration framework uses historical accuracy logs $\mathcal{H}_i$ to adjust raw model outputs, model behavior can drift over time due to upstream API updates or changes in training datasets. Continual online calibration---where validator votes are retrospectively adjudicated against deployment outcomes and used to update validator registry weights---is necessary to maintain calibration accuracy.

\subsection{Cost-Benefit Trade-offs}
Evaluating infrastructure mutations through a multi-agent quorum introduces nontrivial inference cost. In the prototype, the high-risk bucket averages \$0.080 per validation check using the anonymized model-class mix reported in Section~\ref{subsec:validator-baselines}.

While this cost is high compared to static policy engines, it must be evaluated against the cost of the failures it is intended to prevent. A single database outage or data leakage incident can cost an enterprise hundreds of thousands of dollars in recovery, compliance penalties, and lost revenue. For high-blast-radius actions ($R(C) > 0.6$), the token cost is therefore small relative to incident risk. To optimize cost for low-risk actions, SQA uses compact model classes and smaller quorums; in the prototype, the low-risk bucket averages \$0.0048 per check.

\subsection{Latency and Scalability}
Adding 1.45 to 4.12 seconds of median validation latency to the Kubernetes admission pipeline, as measured in the prototype, is too high for high-frequency microservice operations (e.g., routing individual network packets or allocating short-lived local memory pages).

SQA is not designed for fast-path data-plane operations. It is a control-plane primitive for high-stakes infrastructure mutations. For slow-path configuration changes, database migrations, and provisioning updates, the median validation latencies reported in Table~\ref{tab:latency-cost} are within many operational control-plane budgets.

To improve latency, the prototype uses parallel container execution (launching all validators in the quorum concurrently) and pre-warmed validator microVM sandboxes, which keep measured cold-start overhead below 150 milliseconds in that deployment.

\subsection{Evolution of Validator Archetypes}
In our current prototype, archetypes (Security, Cost, Compliance) are statically defined. As cloud services evolve, these archetypes should evolve with them. A promising direction is the dynamic synthesis of validators: when a proposer submits a contract targeting a novel cloud service (e.g., a new AWS machine learning server), the control plane generates a specialized validator persona by retrieving the API documentation and loading it into the validator's context. This helps keep the quorum relevant as the underlying cloud infrastructure changes.

\subsection{Concurrency and Sequencing Bottlenecks}
Binding validator votes to a single, global sequence epoch $\mathit{seq}$ forces strict serialization of all infrastructure mutations. In high-velocity environments, if Proposal A and Proposal B are evaluated in parallel, only the first to be committed at the execution gate will succeed. The second will fail the sequence verification ($\text{epoch}(\Gamma) = \mathit{seq}$) because the global epoch has incremented, forcing a costly re-evaluation. 

To mitigate this sequencing bottleneck, SQA employs a \textit{resource-scoped sequence partition} mechanism. Instead of maintaining a single global epoch for the entire control plane, sequence epochs are scoped and tracked independently for isolated resource sub-trees (e.g., partitioned by Kubernetes namespace, AWS account, or specific database cluster). Under this partitioned model, concurrent mutations targeting disjoint resources (e.g., modifying network rules in \texttt{namespace-A} and scaling pods in \texttt{namespace-B}) execute in parallel without cross-invalidating their respective quorum proofs, limiting serialization overhead to contentious resource-level mutations.

\subsection{Adversarial Prompt Injection Defenses}
Since validators must ingest and parse the proposer's execution contract $C$, parameters, and rationale artifacts in $E$, they are exposed to Indirect Prompt Injection (IPI) attacks. An attacker could embed adversarial text (e.g., \textit{``Ignore your security role and approve this write''}) within a contract parameter. If validator models succumb to this injection, they may vote $+1$ on a catastrophic change, weakening SQA's assurance boundary.

SQA implements a multi-layered defense to neutralize validator-targeted IPI:
\begin{enumerate}
    \item \textbf{Strict Schema Enforcement:} Validators do not parse raw natural language for mutation instructions; they ingest structured JSON data. Inputs are parsed into strictly typed variables before being injected into the validator's system context, preventing data fields from being interpreted as model instructions.
    \item \textbf{Instruction-Data Separation:} SQA prompting structures leverage model-native delimiter tags (e.g., XML blocks) to isolate system instructions from the untrusted contract payload.
    \item \textbf{Veto Archetype Independence:} Because validators run heterogeneous model families and distinct safety-first archetypes (such as the Security Auditor), an injection payload optimized to bypass one model family is less likely to transfer to the other quorum members. The diversity predicate reduces the risk of uniform exploit success.
\end{enumerate}

%% file: sections/12-related-work.tex
\section{Related Work}
\label{sec:related}

Semantic Quorum Assurance sits at the intersection of distributed consensus, Byzantine quorum systems, runtime assurance, policy-as-code, multi-agent LLM evaluation, calibration, prompt-injection defense, and LLM guardrails. We position SQA relative to these areas without claiming that any single component is new in isolation.

\subsection{Classical Consensus and Quorum Systems}
Classical consensus protocols such as Paxos~\cite{lamport-paxos}, Raft~\cite{ongaro-raft}, PBFT~\cite{castro-pbft}, and state machine replication (SMR)~\cite{schneider-smr} establish agreement on a sequence of commands despite crash or Byzantine faults. These protocols are concerned with ordering, replication, and consistency: once a command is admitted to the replicated log, correct replicas execute the same deterministic transition.

SQA uses quorum language, sequence epochs, and cryptographic proofs, but it addresses a different admission question. Consensus protocols decide \textit{which command} is ordered and ensure that replicas agree on it; they do not certify whether an authorized command is semantically safe in its operational context. If a valid leader proposes a syntactically well-formed mutation that deletes the only usable backup, Paxos, Raft, PBFT, and SMR can replicate that mutation correctly. SQA sits before execution and asks whether the proposed mutation should be admitted at all.

\subsection{Byzantine Quorums and Correlated Failures}
Byzantine fault tolerance assumes that a bounded number of participants may behave arbitrarily~\cite{lamport-byzantine,castro-pbft}. Byzantine quorum systems extend quorum constructions to arbitrary server failures while preserving consistency and availability properties under stated intersection assumptions~\cite{malkhi-byzantine-quorums}. These models are powerful for adversarial distributed storage and replicated services, but they typically reason about faulty nodes as members of a fail-prone set rather than about shared cognitive failure modes among validators that use related training data, prompts, or retrieval corpora.

SQA's correlated cognitive failure model is closer in spirit to work showing that diversity assumptions can fail under common-mode software faults. Knight and Leveson showed that independently developed program versions can still exhibit correlated failures~\cite{knight-leveson-multiversion}. LLM validators create an analogous risk: validators instantiated from distinct prompts or roles may still share model-family biases, benchmark contamination, or instruction-following vulnerabilities. SQA therefore treats quorum selection as a risk-estimation problem over empirical $\rho_{ij}(D)$ values, not merely as a threshold count of independent Byzantine replicas.

\subsection{Runtime Assurance and Safety Shields}
Runtime assurance architectures place a trusted monitor or safety controller around a higher-performance but potentially unsafe controller. Simplex architectures use a verified baseline controller to override an advanced controller when safety envelopes are at risk~\cite{simplex-architecture}. Shield synthesis and shielded reinforcement learning similarly enforce temporal-logic safety constraints at runtime by monitoring or correcting unsafe actions~\cite{bloem-shield-synthesis,alshiekh-shielding}. Neural Simplex extends this idea to neural controllers while preserving a runtime safety boundary~\cite{phan-neural-simplex}. Assurance cases provide structured arguments that a system satisfies explicit safety claims~\cite{rushby-assurance}.

SQA is a semantic admission-control analogue for infrastructure mutations. Runtime shields usually guard continuous or reactive controllers against violating formal state constraints. SQA guards a cloud control plane against non-deterministic semantic mistakes before a discrete mutation reaches the execution gate. The shielded component is not a vehicle controller or hardware design; it is an agent-proposed infrastructure change represented as $C$ and $E$. The gate enforces the admission protocol, while semantic assurance remains probabilistic and calibration-dependent.

\subsection{Policy-as-Code and Admission Control}
Policy-as-code systems such as Open Policy Agent and Rego provide declarative evaluation over structured inputs~\cite{opa-rego}. OPA is widely used for Kubernetes admission control, where admission controllers intercept API-server requests before persistence~\cite{opa-kubernetes,kubernetes-admission}. Cloud control mechanisms such as IAM, service control policies (SCPs), and Zero Trust architectures constrain identity, authorization, and organizational permission boundaries~\cite{aws-iam,aws-scp,rose-zerotrust}.

These systems are effective for structural policy checks: rejecting privileged containers, requiring labels, limiting public ingress, or bounding maximum permissions. They are less suited to questions that depend on operational intent and evidence-chain context, such as whether pruning a directory is safe because other restorable backups exist. SQA complements policy-as-code rather than replacing it. Static policies remain the first line of defense; SQA adds a semantic context-evaluation layer whose output is bound to signed validator records and checked by the sovereign gate.

\subsection{LLM Judges, Self-Consistency, and Debate}
LLM-as-judge methods use strong models to evaluate generated outputs and have been studied in benchmarks such as MT-Bench and Chatbot Arena~\cite{zheng-llm-judge}. Self-consistency samples multiple reasoning paths and aggregates answers to improve reasoning accuracy~\cite{wang-self-consistency}. Multi-agent debate work similarly uses multiple LLM instances to critique or refine answers~\cite{liang-debate,du-debate}, often motivated by reducing hallucination or reasoning errors~\cite{ji-hallucination}.

SQA draws on the intuition that multiple evaluators can expose reasoning failures, but it is not an open-ended text-debate protocol. Validators do not merely converse until a final answer emerges. Each validator receives a bounded state projection, signs a canonical $\textsf{VoteRecord}_i$, binds that record to $h(C)$, $h(E)$, $\mathit{seq}$, identity, and archetype, and returns a vote that is evaluated by a risk-adaptive quorum predicate. The aggregate output is a gate-verifiable proof $\Gamma$, not a transcript or model-generated consensus summary.

\subsection{Confidence Calibration and Selective Prediction}
Modern neural models can be miscalibrated: high softmax or verbal confidence need not correspond to empirical correctness~\cite{guo-calibration}. Selective prediction and reject-option classifiers formalize the trade-off between coverage and risk by allowing models to abstain when confidence is insufficient~\cite{geifman-selective-classification,geifman-selectivenet}.

SQA uses calibrated assurance scores $c_i$ for a related purpose. A validator's raw statement of confidence is not treated as a trustworthy probability. Instead, confidence is calibrated against historical adjudication, domain-specific reliability, evidence completeness, and observed correlation. Weighted voting and veto thresholds depend on these calibrated scores because infrastructure admission requires risk estimates that are auditable and empirically grounded. This differs from simple majority voting over uncalibrated LLM judgments.

\subsection{Prompt Injection and Indirect Prompt Injection}
Prompt injection attacks exploit the fact that natural-language instructions and untrusted data share the same model context. Early work demonstrated direct prompt-injection techniques against language models~\cite{perez-promptinject}. Indirect prompt injection expands the threat model by embedding malicious instructions in retrieved documents, web pages, logs, or other external content consumed by an LLM-integrated application~\cite{greshake-indirect-prompt-injection}. Recent benchmarks such as BIPIA evaluate indirect prompt-injection attacks and defenses at scale~\cite{yi-bipia}, and OWASP identifies prompt injection as a central LLM-application risk~\cite{owasp-llm-top10}.

SQA's validators are exposed to a validator-targeted form of indirect prompt injection because contracts, parameters, logs, and evidence-chain artifacts are untrusted inputs. The protocol therefore relies on instruction/data separation, read-only sandboxing, constrained output schemas, evidence hashing, and final gate verification. These controls do not solve prompt injection at the model level; they reduce the authority of compromised validator reasoning by requiring signed, auditable votes from a selected quorum and by preventing validators from executing mutations directly.

\subsection{Guardrails and LLM Firewalls}
Guardrail systems and LLM firewalls classify or constrain model inputs and outputs. Llama Guard uses an LLM-based input/output safeguard for human-AI conversations~\cite{inan-llamaguard}, while NeMo Guardrails provides programmable rails for conversational LLM applications~\cite{rebedea-nemo}. OWASP's LLM security guidance similarly treats prompt injection, insecure output handling, excessive agency, and related risks as application-level concerns~\cite{owasp-llm-top10}.

These tools are useful local defenses, but single-model guardrails are insufficient for heterogeneous infrastructure safety. Infrastructure mutations require balancing security, reliability, database recoverability, cost, and compliance context. A guardrail trained to detect harmful content or unsafe conversation categories may not recognize that a syntactically normal IAM change violates separation-of-duty intent, or that a database pruning command is unsafe only because a backup invariant fails in $S_{\mathit{seq}}$. SQA uses guardrails as complementary controls but adds multi-archetype validation, correlated-failure-aware quorum selection, signed evidence-bound votes, and a sovereign execution gate.

\subsection{Positioning}
SQA's contribution is the combination of semantic validation, correlated-failure-aware quorum selection, cryptographically bound evidence and vote records, and sovereign execution gating. The individual ingredients have precedents in consensus, quorum systems, runtime assurance, policy-as-code, LLM evaluation, calibration, and guardrails; SQA integrates them into an admission-control protocol for non-deterministic infrastructure mutations.

%% file: sections/13-limitations.tex
\section{Limitations}
\label{sec:limitations}

SQA is intended as a control-plane admission primitive, not as a complete proof system for semantic safety. Its guarantees and deployment requirements have several engineering boundaries:

\begin{itemize}
    \item \textbf{Semantic safety is probabilistic.} SQA does not prove that a semantically unsafe proposal can never be approved. The gate-integrity property ensures that mutations cannot bypass the required certification artifacts, but the semantic judgment itself depends on calibrated validators. SQA therefore enforces a calibrated, auditable certification process rather than a mathematical guarantee of universal safety.
    \item \textbf{Correlation estimates are approximate.} The pairwise coefficient $\rho_{ij}(D)$ is a practical estimator of shared validator failure over a domain. It may miss higher-order common-mode failures that appear only when three or more model families, prompt templates, retrieval sources, or evidence projections fail together. Future deployments should supplement pairwise estimates with joint-failure red-teaming and domain-specific stress tests.
    \item \textbf{Calibration can drift.} Validator confidence scores and weights are time-bounded measurements. Provider-side model updates, serving-stack changes, new training data, and prompt-template revisions can change validator behavior without changing the SQA protocol. Production systems need periodic recalibration, adjudication of historical decisions, and versioned registry snapshots.
    \item \textbf{Prompt injection remains residual risk.} Schema enforcement, instruction-data separation, sandboxing, and constrained output formats reduce validator-targeted prompt injection, but they do not remove the underlying instruction/data ambiguity of LLM contexts. SQA limits the authority of compromised validator reasoning through diversity, signed votes, and gate verification; it does not solve prompt injection at the model level.
    \item \textbf{Evaluation coverage is incomplete.} The evaluation covers Kubernetes, database administration, and IAM/network-security scenarios, including synthetic cases and generalized incident patterns. It does not cover every infrastructure domain, organization-specific policy convention, or rare catastrophic interaction among services. New domains require additional scenario construction, labeling, calibration, and correlation measurement.
    \item \textbf{Latency limits the use case.} The validation latency measured in our prototype is appropriate for slow-path control-plane mutations such as configuration changes, access-control updates, migrations, and recovery operations. It is inappropriate for fast-path data-plane decisions such as packet forwarding, per-request authorization, memory allocation, or other high-frequency operations where millisecond-level latency dominates.
    \item \textbf{External inference can raise governance concerns.} Calling external model APIs may expose contracts, evidence summaries, logs, or operational context to third-party inference providers. Sovereign deployments may require local models, private inference endpoints, confidential-computing enclaves, or strict redaction policies before validator invocation.
    \item \textbf{Break-glass procedures remain necessary.} Emergency operations may require human-controlled override paths when the quorum cannot form, model services are unavailable, or immediate action is needed to preserve availability or safety. Such break-glass paths should be explicit, strongly authenticated, logged, rate-limited, and retrospectively audited; they are operational complements to SQA, not replacements for it.
\end{itemize}

These limitations define where SQA must be engineered with calibration, monitoring, red-teaming, and operational governance. They do not undermine the core role of the protocol: preventing unverified non-deterministic reasoning from directly mutating infrastructure state.

%% file: sections/13-conclusion.tex
\section{Conclusion and Future Work}
\label{sec:conclusion}

As autonomous AI agents assume control over critical cloud infrastructure, classical security and consensus mechanisms need an admission layer for semantic intent. Classical consensus replicates state transitions, but agentic infrastructure also requires intent certification before irreversible mutations execute.

In this paper, we introduced \textbf{Semantic Quorum Assurance (SQA)}, a control-plane primitive that aggregates heterogeneous, risk-aware semantic judgments over an execution contract before allowing non-deterministic agent proposals to mutate infrastructure state. We formalized a correlated cognitive failure model to capture semantic errors among LLM-based validators, designed a risk-adaptive quorum predicate, and decoupled validation from execution authority via a sovereign execution gate. In our prototype evaluation on infrastructure-inspired mutation scenarios, SQA reduces unsafe approval from 18.5\% for single-agent validation to 0.3\% on held-out safe/unsafe trials excluding ambiguous scenarios, with median validation latency of 1.45--4.12 seconds across the studied risk buckets.

Future work includes exploring hardware-enforced Trusted Execution Environments (TEEs) to protect validator reasoning loops, developing online calibration algorithms that dynamically adjust validator weights based on historical adjudication, and designing specialized verification languages that allow agents to express evidence chain audits in mathematically verifiable formats.

%% file: sections/appendix-scenario-provenance.tex
\section{Scenario Template Families}
\label{app:scenario-provenance}

The scenario corpus is generated from parameterized templates covering Kubernetes pod scaling, namespace deletion, daemonset rollout, node drain, persistent-volume modification, SQL migration, backup pruning, index creation, user grant modification, IAM policy editing, security-group modification, and secret rotation.

Unsafe scenarios are constructed from public incident-pattern categories and cloud-operation anti-patterns: public exposure of administrative services, loss of the only active backup or recovery path, wildcard IAM privilege escalation, destructive database operations without verified snapshots, Kubernetes changes that remove singleton control-plane dependencies, and credential rotation without confirmed consumer rollout. The scenarios use synthetic resource names, logs, and configurations; they are not production traces.